\documentclass[conference,a4paper]{IEEEtran}
\IEEEoverridecommandlockouts

\usepackage[hidelinks]{hyperref}
\usepackage[cmex10]{amsmath}
\usepackage{amssymb,amsfonts}
\usepackage{dblfloatfix}

\usepackage[ruled,vlined]{algorithm2e}
\usepackage{graphicx}
\graphicspath{{Figures/PDF/}{Figures/PNG/}}

\usepackage{booktabs}
\usepackage{siunitx}
\usepackage[numbers,compress]{natbib}
\usepackage{texnames}
\usepackage{bm,bbm}
\usepackage{orcidlink}

\begin{document}

\title{GRIT: Graph Transformer For Internal Ice Layer Thickness Prediction
\thanks{This work is supported by NSF BIGDATA awards (IIS-1838230, IIS-2308649), NSF Leadership Class Computing awards (OAC-2139536), IBM, and Amazon.}
\thanks{
© 2025 IEEE. Published in the 2025 IEEE International Geoscience and Remote Sensing Symposium (IGARSS 2025), scheduled for 3 – 8 August 2025 in Brisbane, Australia. Personal use of this material is permitted. However, permission to reprint/republish this material for advertising or promotional purposes or for creating new collective works for resale or redistribution to servers or lists, or to reuse any copyrighted component of this work in other works, must be obtained from the IEEE. Contact: Manager, Copyrights and Permissions / IEEE Service Center / 445 Hoes Lane / P.O. Box 1331 / Piscataway, NJ 08855-1331, USA. Telephone: + Intl. 908-562-3966.}
\thanks{This version is the accepted manuscript submitted to arXiv. The final version will be published in the Proceedings of IGARSS 2025 and available via IEEE Xplore. For citation, please refer to the published version in IGARSS 2025.
}
}

\author{\IEEEauthorblockN{Zesheng Liu}
\IEEEauthorblockA{\textit{Department of Computer Science and Engineering} \\
\textit{Lehigh University}\\
Bethlehem, USA \\
zel220@lehigh.edu}
\and
\IEEEauthorblockN{Maryam Rahnemoonfar\IEEEauthorrefmark{1}}
\IEEEauthorblockA{\textit{Department of Computer Science and Engineering} \\
\textit{Department of Civil and Environmental Engineering}\\
\textit{Lehigh University}\\
Bethlehem, USA \\
maryam@lehigh.edu}
\thanks{\IEEEauthorrefmark{1} Correspondence to maryam@lehigh.edu}
}

\maketitle

\begin{abstract}
Gaining a deeper understanding of the thickness and variability of internal ice layers in Radar imagery is essential in monitoring the snow accumulation, better evaluating ice dynamics processes, and minimizing uncertainties in climate models. Radar sensors, capable of penetrating ice, capture detailed radargram images of internal ice layers. In this work, we introduce GRIT, graph transformer for ice layer thickness. GRIT integrates an inductive geometric graph learning framework with an attention mechanism, designed to map the relationships between shallow and deeper ice layers. Compared to baseline graph neural networks, GRIT demonstrates consistently lower prediction errors. These results highlight the attention mechanism's effectiveness in capturing temporal changes across ice layers, while the graph transformer combines the strengths of transformers for learning long-range dependencies with graph neural networks for capturing spatial patterns, enabling robust modeling of complex spatiotemporal dynamics.
\end{abstract}

\begin{IEEEkeywords}
	Deep Learning, Remote Sensing, Graph Transformer, Ice Layer, Ice Thickness.
\end{IEEEkeywords}

\section{Introduction}

Global warming, climate change, and sea level rise are not abstract concerns—they are urgent challenges that affect all aspects of life on One Earth. Polar ice sheets hold vital clues about both the past and present of our climate system, offering critical insights into the processes driving these global phenomena. These ice sheets, composed of multiple layers formed in different years, preserve a year-by-year record of Earth's climate history, providing a detailed archive for understanding the evolution of our planet's environment. Studying the internal layers of polar ice is essential for monitoring snow accumulation, improving the tracking of sea level changes and their impacts on global ecosystems, and reducing uncertainties in climate models. By leveraging advancements in remote sensing technologies, we can unravel the thickness and variability of these layers, empowering researchers to take informed actions to mitigate climate impacts. 

The traditional method to study the internal ice layer is through onsite ice cores. This method involves drilling holes at various locations across the polar ice sheets to manually extract large cylindrical ice samples. However, this approach has significant limitations. The coverage provided by ice cores is sparse and discontinuous, making it difficult to develop a comprehensive understanding of how ice layers vary across regions. Moreover, since the approach is onsite measurement and involves some manual process, sampling is restricted to locations that can be physically accessed by researchers. While some successes have been achieved by interpolating these scattered measurements, the resulting uncertainties can adversely impact the accuracy of other climate models. Furthermore, onsite ice core extraction is not environmentally friendly and causes physical damage to the ice sheet.

In recent years, airborne radar sensors\cite{AirborneRadar} have become a popular alternative for studying ice layers due to their ability to traverse thick ice and provide continuous measurements. One notable example is the snow radar sensor\cite{snowradar} operated by the Center for Remote Sensing of Ice Sheets (CReSIS). Mounted on aircraft following various flight routes, airborne snow radar sensor transmits signals that penetrate thick ice sheets, capturing the internal ice layers by analyzing the strength of the reflected signals\cite{Arnold_2020}. Radargram images, as shown in Figure \ref{fig:diagram} (\textbf{a}), are the resulting data of the snow radar sensor that depicts internal layers at various depths.

Recently, graph convolutional networks have achieved great success in various aspects, including the prediction of deep ice layer thickness. Zalatan et al.\cite{zalatan_icip,Zalatan_igarss,Zalatan2023}, liu et al.\cite{liu2024learningspatiotemporalpatternspolar,liu2024multibranchspatiotemporalgraphneural} and Rahnemoonfar et al.\cite{PI_GCNLSTM} proposed representing internal ice layers as spatial graphs and using different kinds of spatio-temporal graph neural networks to model the relationship between layers across different years. While GNNs provide a promising framework for modeling spatiotemporal patterns of ice layers, they are inherently limited in their ability to capture global dependencies due to their reliance on localized feature aggregation operations. This limitation can hinder the network's performance in learning complex temporal changes, which usually requires the understanding of global temporal relationships.

In this paper, we address the limitation of graph neural networks by introducing a novel graph transformer network, named \textbf{G}raph t\textbf{R}ansfromer for \textbf{I}ce layer \textbf{T}hickness (GRIT). GRIT integrates the attention mechanism into the geometric deep learning framework. Unlike traditional GNNs, GRIT leverages attention to dynamically model both local and global dependencies, offering greater flexibility and robustness in capturing complex temporal patterns. Our major contributions are:
\begin{itemize}
    \item We developed GRIT, a graph transformer network that can learn the spatiotemporal patterns from the upper $m$ ice layers and make predictions for the underlying $n$ layers.
    \item Our proposed network uses GraphSAGE to capture spatial patterns as feature embeddings and uses a temporal attention block to effectively learn both long-term dependencies and short-term dynamics.
    \item We conduct extensive experiments to compare our proposed network with recurrent graph convolutional networks as baselines, and the results show that our proposed network, GRIT, can consistently have a lower root mean squared error.
\end{itemize}

\section{Related Work}

\subsection{Ice Layer Boundary Tracking}
Identifying the ice layer boundaries from radargram images is a complex task due to the fact that deep ice layers formed a long time ago may be broken, incomplete, or even melted \cite{LearnSnowLayerThickness}. Deep learning methods, especially convolutional neural networks (CNNs) and generative adversarial networks (GANs), have been developed to accurately track the ice layers from the radargram images \cite{DeepIceLayerTracking,DeepLearningOnAirborneRadar,Rahnemoonfar_2021_JOG,Yari_2021_JSTAR}. Although reaching some success, all these methods highlight that noise in the input radargram, together with the lack of high-quality snow radar datasets and annotations, is currently the main obstacle. Some researchers also adapt the idea of physics-informed learning into the task of detecting ice layer boundaries, where physical laws are introduced to better denoising the input radargram images\cite{DeepHybridWavelet,varshney2021refining} or pretraining the neural network\cite{LearnSnowLayerThickness}. 

Unlike previous convolution-based neural network, our proposed GRIT network focus on geometric deep learning and attention mechanism, demonstrating enhanced robustness to noise and delivering more stable and reliable performance across varying quality of input radargrams.


\subsection{Graph Neural Network For Ice Layer Thickness Prediction}

Zalatan et al.\cite{zalatan_icip,Zalatan_igarss,Zalatan2023} developed a multi-target, adaptive long short-term memory graph convolutional network (AGCN-LSTM), to predict the thickness of deeper internal ice layers. By plugging graph convolutional network (GCN) into long short-term memory (LSTM), the network can learn the spatial patterns within each internal layer and temporal changes across different layers simultaneously. EvolveGCNH\cite{EGCN} is used as an adaptive layer to improve the model's learning ability for more complicated features and the model's robustness to noisy inputs. Their idea is further extended by liu et al.\cite{liu2024learningspatiotemporalpatternspolar, liu2024multibranchspatiotemporalgraphneural} and Rahnemoonfar et al.\cite{PI_GCNLSTM} to improve both the accuracy and efficiency by using a multi-branch structure or introducing data from the physical model. 

Compared with their work, our proposed GRIT network uses a temporal attention block to learn the temporal dependencies more effectively, capturing both short-term and long-term patterns across internal ice layers.

\section{Radargram Dataset}

In this paper, our radargram dataset is captured across the Greenland region in 2012, via an airborne snow radar sensor operated by CRsSIS\cite{CReSIS_radar}, as part of the NASA's Operation Ice Bridge\cite{Leuschen2011SnowRadar}. Airborne snow radar sensor is currently the most effective way to measure the status of internal ice layers, as radar sensor is the main sensor that can penetrate through the thick ice sheet. The internal ice layers, captured by the airborne snow radar sensor as radargram images, are shown in Figure \ref{fig:diagram}(\textbf{a}), where pixel values represent the strength of the reflected signal, with brighter pixels indicating stronger reflections\cite{Arnold_2020}. Labeled images are then generated by manually labeling the boundaries of each ice layer in the radargram by NASA scientists, as illustrated in Figure \ref{fig:diagram}(\textbf{b}). Based on the labeled image, the thickness of the ice layer at each pixel is computed as the difference between the value of upper and lower boundaries. Furthermore, during radargram acquisition, latitude and longitude coordinates for each pixel are recorded simultaneously by the airborne snow radar sensor. 

\section{Key Designs}
\begin{figure*}[!t]
  \centering
  \includegraphics[width=0.9\textwidth]{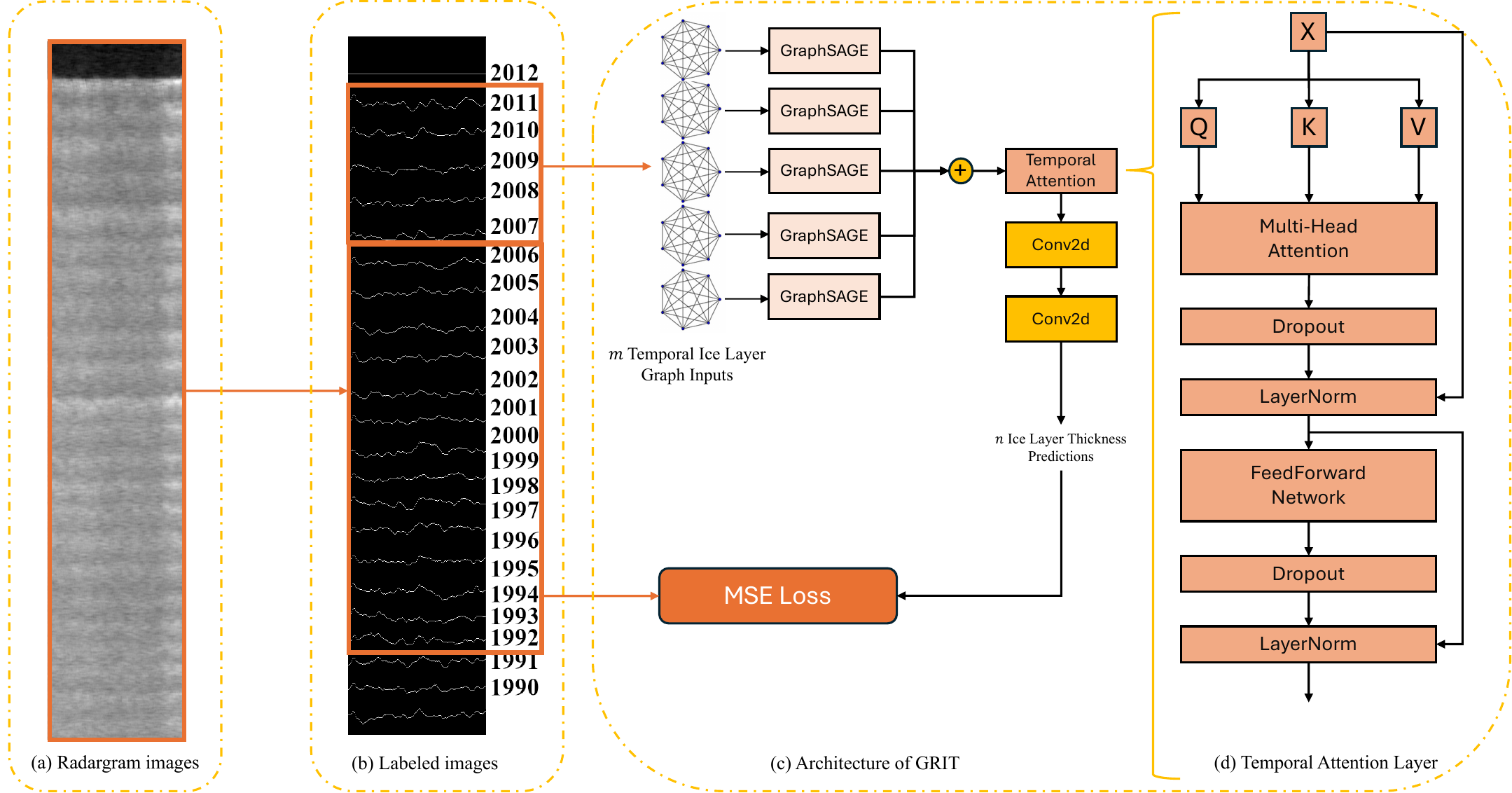}
  \caption{Diagram of our dataset and our proposed graph transformer network. (\textbf{a}) Radargram images (\textbf{b}) Labeled image, where the boundaries of each ice layer are manually labeled out. (\textbf{c}) Architecture of GRIT, our proposed graph transformer network for ice layer thickness prediction. (\textbf{d}) Temporal attention layer in GRIT.\label{fig:diagram}}
\end{figure*}

We propose GRIT, a graph transformer network built to learn the spatiotemporal patterns of the top $m$ internal ice layers and make predictions for the $n$ underlying layer. The primary motivation is to enhance the network's ability to capture complex temporal changes via self-attention mechanism and transformer network. Unlike previous CNN, RNN, or GNN-based methods that primarily focus on local patterns and may suffer from issues like vanishing gradients, over-smoothing, or limited receptive fields, the self-attention mechanism and transformer networks can capture long-range dependencies regardless of sequence length. Additionally, they are more robust to noise by dynamically assigning attention weights, allowing the model to focus on the most relevant parts of the input while ignoring irrelevant or noisy information.

The input of our network is temporal sequences of $m$ spatial graph. Figure \ref{fig:diagram}(\textbf{c}) shows the overall structure of GRIT, including a GNN part designed to learn the spatial patterns within each spatial graph and concatenate the output as feature embeddings. The learned embeddings are then passed through a temporal attention encoder to learn the temporal changes across different graphs. Two convolution layers are used as the decoder to make the final prediction.

\subsection{GraphSAGE Inductive Framework}

We used five independent GraphSAGE to capture the spatial patterns within each internal layer and form the feature embeddings. Compared with Graph Convolution Network\cite{kipf2017_GCN}, GraphSAGE \cite{hamilton2018inductive_graphsage} is an inductive framework designed to create node embeddings for previously unseen data by leveraging local neighbor sampling and feature aggregation \cite{ZHOU202057_Review}. For a given node $i$ and its input node feature $x_i$, the learned feature embedding via GraphSAGE $\textbf{x}'_i$ is defined as: 
\begin{equation}
\textbf{x}'_i = \textbf{W}_1 \textbf{x}_i + \textbf{W}_2 \cdot \text{mean}_{j \in \mathcal{N}(i)} \textbf{x}_j
\end{equation}

where $\textbf{W}_1,\textbf{W}_2$ are learnable layer weights, $N(i)$ is the neighbor list of node $i$ that includes neighbors with different depth, $\textbf{x}_j$ is the neighbor node features and mean is the aggregation function. In GraphSAGE, it separates the transformation of root and neighbor nodes by having separate $\textbf{W}_1,\textbf{W}_2$ matrix. This separation ensures that each node can better preserve its own characteristics and reduce the effect of outlier neighbor nodes, resulting in more robust and efficient learning.

\subsection{Temporal Attention Layer}
In the encoder of GRIT, we adapted the standard multi-head attention proposed by Vaswani et al.\cite{vaswani2017attention}, which stacks a few scaled dot-product attention together for better performance. For each input feature embedding matrix $X$, we generate the queries, keys, and values for each head, defined as:
\begin{equation}
    Q_i=XW_i^Q, K_i=XW_i^K, V_i=XW_i^V
\end{equation}
Each head then performs the attention calculation independently, using its own query, key, and value:
\begin{equation}
    Attention(Q_i,K_i,V_i)=softmax(\frac{Q_iK_i^T}{\sqrt{d_k}})V_i
\end{equation}
where $d_k$ refers to the dimension of the key vector in each head. The outputs of each head are finally concatenated together to produce the output of the multi-head attention:
\begin{equation}
    MHA(X) = Concat(Head_1, Head_2,...,Head_n)W^o
\end{equation}
where $W^o$ is a learnable weight matrix that combines the results of all the heads, and $Head_i = Attention(Q_i, K_i, V_i)$. As shown in Figure \ref{fig:diagram}(\textbf{d}), our temporal attention block also contains dropout, layer normalization, feedforward network, and skip connections, which are common components in an attention encoder. In the temporal attention block of GRIT, we use 8 heads in total, same as number of heads used by Vaswani et al.\cite{vaswani2017attention}. Additionally, considering the fact that our feature embedding matrix $X$ will have both a spatial dimension and a temporal dimension, necessary transpose operations are performed to ensure the temporal attention encoder is applied to the temporal dimension.

\section{Experiements and Results}

\begin{table}[!t]
    \caption{Experiment results of GCN-LSTM, GraphSAGE-LSTM, Multi-branch graph neural network, and proposed GRIT.}
    \begin{center}
    \begin{tabular}{ccc}
    \toprule
\textbf{Model}                 & \textbf{RMSE} \\ 
\midrule
AGCN-LSTM\cite{zalatan_icip}             & 3.4808 $\pm$ 0.0397  \\
GCN-LSTM\cite{Zalatan2023}              & 3.1745 $\pm$ 0.1045   \\
GraphSAGE-LSTM\cite{liu2024learningspatiotemporalpatternspolar}        & 3.3837 $\pm$ 0.1103 \\
Multi-branch\cite{liu2024multibranchspatiotemporalgraphneural} & 3.1087 $\pm$ 0.0555 \\
GRIT(Ours) & \textbf{3.0597 $\pm$ 0.0326}\\
\bottomrule
\end{tabular}
\end{center}
    \label{table:OverallResults}
\end{table}

\subsection{Data Preprocessing and Graph Generation}
Our proposed graph transformer is tested under a specific case where $m=5$ and $n=15$, aiming to use the thickness and geographical information from the top 5 layers (2007-2011) to predict the thickness of the underlying 15 layers (1992-2016). These $m$ and $n$ can be further extended to any number of layers or layers measured by different sensors. Due to variations in snow accumulation, melting, and ice sheet topography, the number of internal layers varies by location. To maintain high data quality, radargram images with fewer than 20 complete internal layers are excluded during preprocessing. Image may be eliminated due to the insufficient number of layers or incompleteness of the top 20 internal ice layers. After data preprocessing, there are 1660 high-quality radargram images in total. These images are then divided into training, validation, and test sets with a ratio of $3:1:1$.


Our graph dataset is generated by converting each ground-truth radargram image into a temporal sequence of five spatial graphs, where each spatial graph represents a single ice layer formed between 2007-2011. Each spatial graph consists of 256 nodes connected by undirected edges. Edge weights are computed as the inverse distance of the geographic distance between node locations via the haversine formula, defined as:
\begin{equation}
    w_{i,j}  = \frac{1}{2\arcsin{(hav(\phi_j-\phi_i)+\cos{\phi_i}\cos{\phi_j}hav(\lambda_j-\lambda_i))}}
\end{equation}
where $i, j$ can be any node in the graph, $w_{i,j}$ is the computed edge weights between nodes, $\phi, \lambda$ are the latitude and longitude coordinates and $hav(\theta) = \sin^2{(\frac{\theta}{2})}$. For each node in the spatial graphs, they will contain three features: latitude, longitude, and thickness.

\subsection{Experiment Details}
To highlight the performance of GRIT, we compared it with several GNN-based methods as baselines, including AGCN-LSTM\cite{zalatan_icip}, GCN-LSTM\cite{Zalatan2023}, SAGE-LSTM\cite{liu2024learningspatiotemporalpatternspolar} and multi-branch spatio-temporal graph neural network\cite{liu2024multibranchspatiotemporalgraphneural}. All the networks were trained on the same machine with 8 NVIDIA A5000 GPUs and Intel(R) Xeon(R) Gold 6430 CPU. Mean-squared error (MSE) loss was chosen as the loss function for all the networks. We used the Adam optimizer with 0.0001 as the weight decay coefficient. For all the baseline graph neural networks, we set the initial learning rate to be 0.01 and used a step learning rate scheduler to half the learning rate every 75 epochs. For GRIT, We used an adaptive learning rate scheduler that adjusts the learning rate based on the average validation loss, starting at 0.001 and halving it to respond to performance stagnation after 16 epochs without improvement. Compared to a step scheduler, this approach dynamically adjusts based on validation performance rather than fixed intervals. To ensure fully convergence, all the networks were trained for 450 epochs. Both GraphSAGE and the adaptive learning rate scheduler may cause some randomness to the model's performance. To better reduce the impact, we create five different versions of the training, validation, and testing datasets by applying various random permutations to all 1,660 valid images before splitting. Each graph neural network was then trained using all five versions of these datasets.

\begin{figure}[!t]
  \centering
  \includegraphics[width=0.35\textwidth]{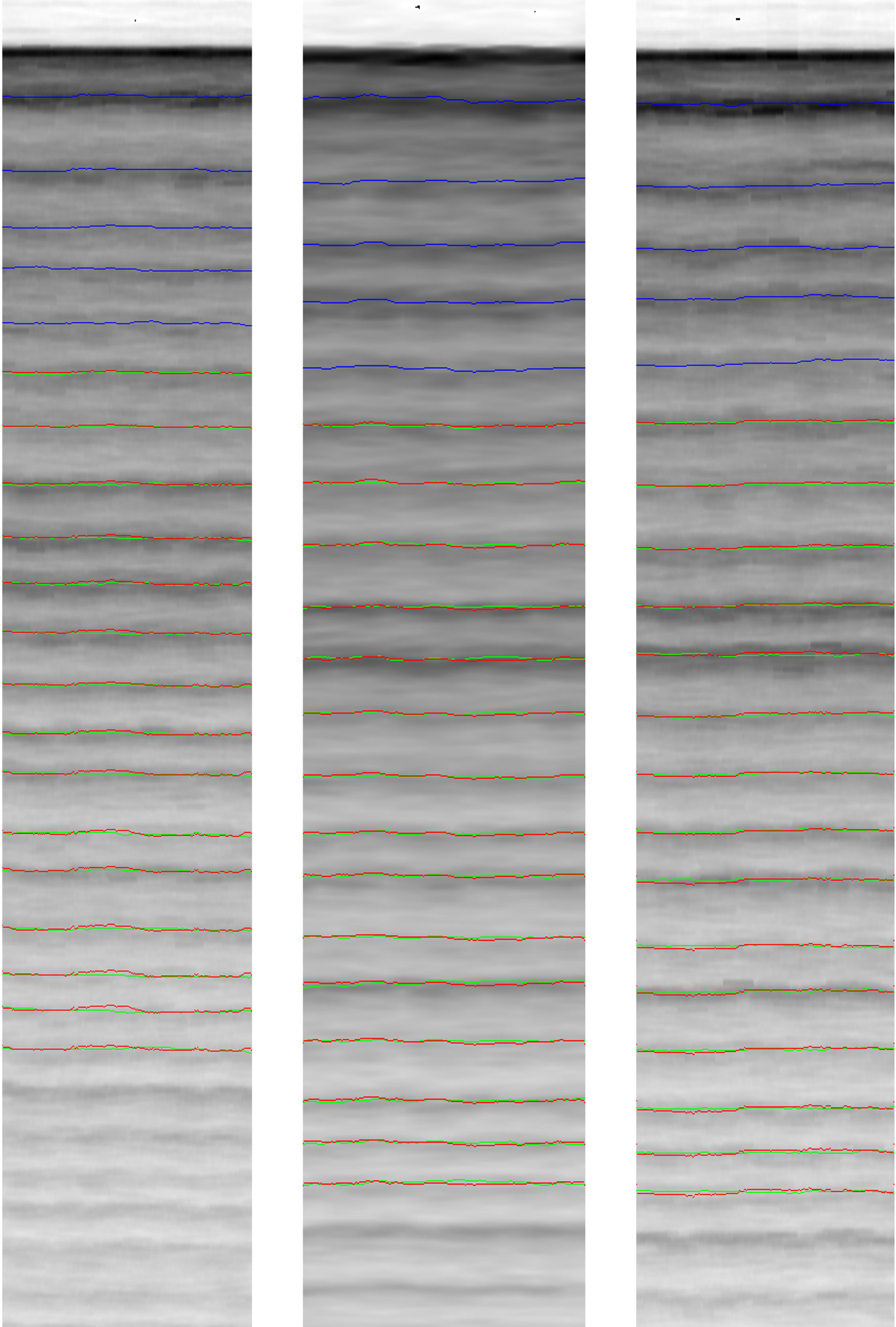}
  \caption{Qualitative results of GRIT predictions. The green line is the groundtruth (manually-labeled ice layers) and the red line is the GRIT prediction.\label{fig:qualitative}}
\end{figure}

\subsection{Results and Analysis}
For each version of the training, validation, and testing datasets, we recorded the root mean squared error (RMSE) between the predicted layer thickness and the ground truth thickness of deeper layer. Then, we calculated the mean and standard deviation of RMSE across the five versions, reported in Table \ref{table:OverallResults}. We can see that by enhancing the model's ability to learn long-range temporal dependencies via a temporal attention block, our proposed GRIT network outperforms previous GNN-based methods with a lower average RMSE and a lower standard deviation. Figure \ref{fig:qualitative} shows the qualitative results of GRIT predictions.

\section{Conclusion}

In this work, we proposed GRIT, a graph transformer designed to predict the thickness of deeper internal ice layers of the Greenland ice sheet, utilizing the thicknesses and geographical information of the shallow ice layers as input. GRIT captures spatial patterns using GraphSAGE for feature embedding and models long-range temporal dependencies through a temporal attention block. We evaluated GRIT on a specific case, using data from ice layers formed between 2007 and 2011 to predict the thickness of layers formed from 1992 to 2006. Notably, GRIT is generalizable to accommodate any number of ice layers and radargrams of varying sizes. Experiments show that our proposed graph transformer, GRIT, performed consistently better than all the baseline graph neural networks.

\small
\bibliographystyle{IEEEtranN}
\bibliography{references}

\end{document}